\documentclass[runningheads]{llncs}
\usepackage{graphicx}
\usepackage{comment}
\usepackage{amsmath,amssymb} % define this before the line numbering.
\usepackage{color}

\usepackage{array}
\usepackage{lipsum}
\def\onedot{.}
\def\eg{\emph{e.g}\onedot} 
\def\ie{\emph{i.e}\onedot} 
 
\def\etc{\emph{etc}\onedot} 
 
\def\etal{\emph{et al}\onedot}
\makeatletter\renewcommand\paragraph{\@startsection{paragraph}{4}{\z@}
  {.5em \@plus1ex \@minus.2ex}{-.5em}{\normalfont\normalsize\bfseries}}\makeatother

\begin{document}
% \linenumbers
\pagestyle{headings}
\mainmatter
\def\ECCVSubNumber{1456}  % Insert your submission number here

\title{Regional Homogeneity: Towards Learning Transferable Universal Adversarial Perturbations Against Defenses} % Replace with your title

\begin{comment}
\titlerunning{ECCV-20 submission ID \ECCVSubNumber} 
\authorrunning{ECCV-20 submission ID \ECCVSubNumber} 
\author{Anonymous ECCV submission}
\institute{Paper ID \ECCVSubNumber}
\end{comment}
%******************

% \begin{comment}
\titlerunning{Regional Homogeneity}
\author{Yingwei Li\inst{1} \and
Song Bai\inst{2} \and
Cihang Xie\inst{1} \and
Zhenyu Liao\inst{3} \and \\
Xiaohui Shen\inst{4} \and
Alan Yuille\inst{1}
}
\authorrunning{Y. Li et al.}
\institute{Johns Hopkins University \and University of Oxford \and Kuaishou Technology \and ByteDance Research}
% \end{comment}
%******************
\maketitle

\begin{abstract}
This paper focuses on learning transferable adversarial examples specifically against defense models (models to defense adversarial attacks). In particular, we show that a simple universal perturbation can fool a series of state-of-the-art defenses.

Adversarial examples generated by existing attacks are generally hard to transfer to defense models. We observe the property of regional homogeneity in adversarial perturbations and suggest that the defenses are less robust to regionally homogeneous perturbations. Therefore, we propose an effective transforming paradigm and a customized gradient transformer module to transform existing perturbations into regionally homogeneous ones. Without explicitly forcing the perturbations to be universal, we observe that a well-trained gradient transformer module tends to output input-independent gradients (hence universal) benefiting from the under-fitting phenomenon. Thorough experiments demonstrate that our work significantly outperforms the prior art attacking algorithms (either image-dependent or universal ones) by an average improvement of~~$14.0\%$ when attacking $9$ defenses in the transfer-based attack setting. In addition to the cross-model transferability, we also verify that regionally homogeneous perturbations can well transfer across different vision tasks (attacking with the semantic segmentation task and testing on the object detection task). The code is available here: \url{https://github.com/LiYingwei/Regional-Homogeneity}.
\end{abstract}

\section{Introduction} \label{sec:intro}
Deep neural networks are demonstrated vulnerable to adversarial examples~\cite{szegedy2013intriguing}, crafted by adding imperceptible perturbations to clean images. The variants of adversarial attacks~\cite{bai2019adversarial,cao2019adversarial,chen2017targeted,eykholt2018robust,gao2020patch,huang2020universal,jia2018poisoning,jin2020adversarial,nitin2018practical,qiu2019semanticadv,sun2018data,sun2020adversarial,zhang2020adversarial} cast a security threat when deploying machine learning systems. To mitigate this, large efforts have been devoted to adversarial defense~\cite{borkar2020defending,jin2020graph,ma2018characterizing,Xiao2020Enhancing}, via adversarial training~\cite{liao2018defense,madry2017towards,tramer2017ensemble,xie2018feature,tang2020transferring,Xie2020intriguing,xie2020sat}, randomized transformation~\cite{das2018shield,guo2017countering,liu2018towards,xie2017mitigating}~\etc.

The focus of this work is to attack defense models, especially in the transfer-based attack setting where models' architectures and parameters remain unknown to attackers. In this case, the adversarial examples generated for one model, which possess the property of ``transferability", may also be misclassified by other models. To the best of our knowledge, learning transferable adversarial examples for attacking defense models is still an open problem.
%which has not been fully studied.

\begin{figure}[tb]
  \begin{minipage}[c]{0.625\textwidth}
    \centering\includegraphics[width=0.95\linewidth]{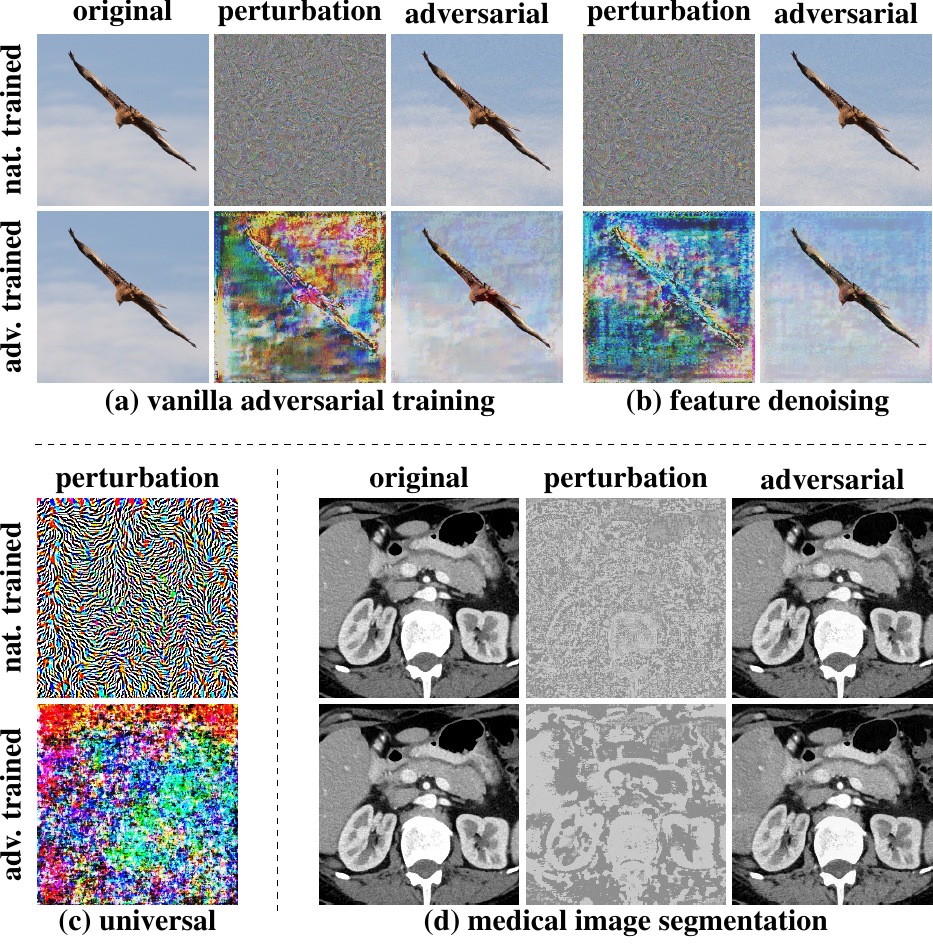}
  \end{minipage}\hfill
  \begin{minipage}[c]{0.325\textwidth}
    % \vspace{-1em}
    \caption{Illustration of region homogeneity property of adversarial perturbations by white-box attacking naturally trained models (top row) and adversarially trained models (bottom row). The adversarially trained models are acquired by (a) vanilla adversarial training~\cite{madry2017towards,xie2018feature}, (b) adversarial training with feature denoising~\cite{xie2018feature}, (c) universal adversarial training~\cite{shafahi2018universal}, and (d) adversarial training for medical image segmentation~\cite{Li2019}} \label{fig:fig2}
  \end{minipage}
%   \vspace{-1em}
\end{figure}

Our work stems from the observation of \textit{regional homogeneity} on adversarial perturbations in the white-box setting. As Figure~\ref{fig:fig2}(a) shows, we plot the adversarial perturbations generated by attacking a naturally trained Resnet-152~\cite{he2016deep} model (top) and an representative defense one (\ie,~an adversarially trained model~\cite{madry2017towards,xie2018feature}). It suggests that the patterns of two kinds of perturbations are visually different. Concretely, the perturbations of defense models reveal a coarser level of granularity, and are more locally correlated and more structured than that of the naturally trained model. The observation also holds when attacking different defense models (\eg,~adversarial training with feature denoising~\cite{xie2018feature}, Figure~\ref{fig:fig2}(b)), generating different types of adversarial examples (image-dependent or universal perturbations~\cite{shafahi2018universal}, Figure~\ref{fig:fig2}(c)), or tested on different data domains (CT scans~\cite{roth2015deeporgan}, Figure~\ref{fig:fig2}(d)).

\begin{figure}[tb]
  % Your images
  \vspace{12pt}
  \includegraphics[width=0.555\linewidth]{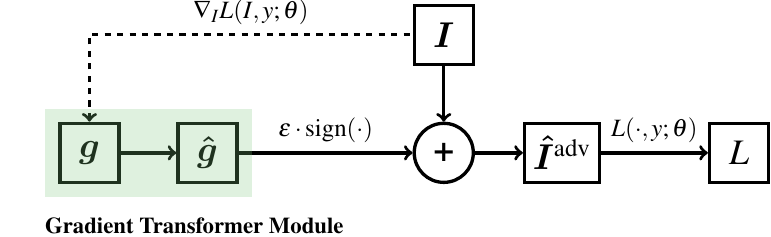} 

  % The caption
  \vspace*{\dimexpr-\parskip-75pt\relax}% Skip backwards over last left-aligned image
  \parshape 8 % Set flow of caption: 6 lines...
    .58\textwidth .42\textwidth % First 5 start @ .5\textwidth with
    .58\textwidth .42\textwidth % a width of .5\textwidth
    .58\textwidth .42\textwidth
    .58\textwidth .42\textwidth
    .58\textwidth .42\textwidth
    .58\textwidth .42\textwidth
    .58\textwidth .42\textwidth
    .58\textwidth .42\textwidth
    % 0pt \textwidth % last (sixth) line restores regular flow ad infinitum
  \makeatletter
  % Setting of actual caption (this is taken from latex.ltx)
  \refstepcounter\@captype\label{fig:train1}% Increase float/caption counter
  \addcontentsline{\csname ext@\@captype\endcsname}{\@captype}% Add content to "List of..."
    {\protect\numberline{\csname the\@captype\endcsname}{ToC entry}}%
  \textbf{\csname fnum@\@captype\endcsname:} % Float caption + #
  \makeatother
  % Actual caption
   Illustration of the transforming paradigm, where $I$ is an original image with the corresponding label $y$, and the gradient $g$ is computed from the naturally trained model $\theta$. Our work learns a mapping to transform gradient from $g$ to $\hat{g}$
  \vspace{0.5em}
\end{figure}

\begin{figure}[tb]
  % Your images
  \hspace{0.49em}\includegraphics[width=0.542\linewidth]{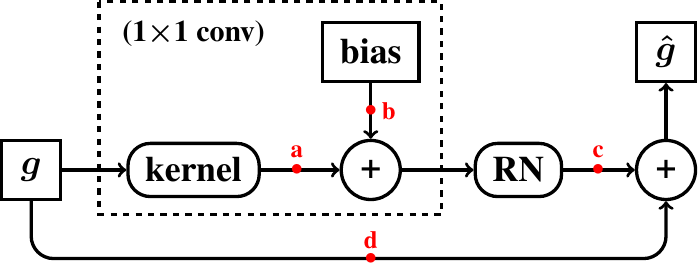} 

  % The caption
  \vspace*{\dimexpr-\parskip-85pt\relax}% Skip backwards over last left-aligned image
  \parshape 9 % Set flow of caption: 6 lines...
    .58\textwidth .42\textwidth % First 5 start @ .5\textwidth with
    .58\textwidth .42\textwidth % a width of .5\textwidth
    .58\textwidth .42\textwidth
    .58\textwidth .42\textwidth
    .58\textwidth .42\textwidth
    .58\textwidth .42\textwidth
    .58\textwidth .42\textwidth
    .58\textwidth .42\textwidth
    0pt \textwidth % last (sixth) line restores regular flow ad infinitum
  \makeatletter
  % Setting of actual caption (this is taken from latex.ltx)
  \refstepcounter\@captype\label{fig:train2}% Increase float/caption counter
  \addcontentsline{\csname ext@\@captype\endcsname}{\@captype}% Add content to "List of..."
    {\protect\numberline{\csname the\@captype\endcsname}{ToC entry}}%
  \textbf{\csname fnum@\@captype\endcsname:} % Float caption + #
  \makeatother
  % Actual caption
  Structure of the gradient transformer module, which has a newly proposed Region Norm (RN) layer, $1 \times 1$ convolutional layer (bias enabled) and identity mapping. We insert four probes (\textcolor{red}{a}, \textcolor{red}{b}, \textcolor{red}{c} and \textcolor{red}{d}) to assist analysis in Section~\ref{sec:universal_analysis} and Section~\ref{sec:universal_exp}
%   \vspace{-1em}
\end{figure}

Motivated by this observation, we suggest that \textit{regionally homogeneous perturbations} are strong in attacking defense models, which is especially helpful to learn transferable adversarial examples in the transfer-based attack setting. Hence, we propose to transform the existing perturbations (those derive from differentiating naturally trained models) to the regionally homogeneous ones. To this end, we develop a novel transforming paradigm (Figure~\ref{fig:train1}) to craft regionally homogeneous perturbations, and accordingly a gradient transformer module (Figure~\ref{fig:train2}), to encourage local correlations within the pre-defined regions. 

The proposed gradient transformer module is quite light-weight, with only $12+2K$ trainable parameters in total, where a $3\times 3$ convolutional layer (bias enabled) incurs $12$ parameters and $K$ is the number of region partitions. According to our experiments, it leads to under-fitting (large bias and small variance) if the module is trained with a large number of images. In general vision tasks, an under-fitting model is undesirable. However in our case, once the gradient transformer module becomes quasi-input-independent (\ie,~aforementioned large bias and small variance), it will output a nearly fixed pattern whatever the input is. Then, our work is endowed with a desirable property,~\ie,~seemingly training to generate image-dependent perturbations, yet get the universal ones. We note our mechanism is different from other universal adversarial generations~\cite{moosavi2016universal,poursaeed2017generative} as we do not explicitly force the perturbation to be universal.

Comprehensive experiments are conducted to verify the effectiveness of the proposed regionally homogeneous perturbation (RHP). Under the transfer-based attack setting, RHP successfully attacks 9 latest defenses~\cite{guo2017countering,kannan2018adversarial,liao2018defense,madry2017towards,tramer2017ensemble,xie2017mitigating,xie2018feature} and improves the top-1 error rates by $21.6\%$ in average, where three of them are the top submissions in the NeurIPS 2017 defense competition~\cite{kurakin2018adversarial} and the Competition on Adversarial Attacks and Defenses 2018. Compared with the state-of-the-art attack methods, RHP not only outperforms universal adversarial perturbations (\eg,~UAP~\cite{moosavi2016universal} by $19.2\%$ and GAP~\cite{poursaeed2017generative} by $15.6\%$), but also outperforms image-dependent perturbations (FGSM~\cite{goodfellow2014explaining} by $12.9\%$, MIM~\cite{dong2017boosting} by $12.6\%$ and DIM~\cite{dong2017boosting,xie2018improving} by $9.58\%$). The achievement over image-dependent perturbations is especially valuable as it is known that image-dependent perturbations generally perform better as they utilized information from the original images. Since it is universal, RHP is more general (natural noises are not related to the target image), more efficient (without additional computational power), and more flexible (\eg, without knowing the target image, people can stick a pattern on the lens to attack artificial intelligence surveillance cameras).

Moreover, we also evaluate the cross-task transferability of RHP and demonstrate that RHP generalizes well in cross-task attack,~\ie,~attacking with the semantic segmentation task and testing on the object detection task.

\section{Related Work}

\paragraph{Transfer-based attacks.}~Practically, attackers cannot easily access the internal information of target models (including its architecture, parameters and outputs). A typical solution is to generate adversarial examples with strong transferability. Szegedy~\etal~\cite{szegedy2013intriguing} first discuss the transferability of adversarial examples that the same input can successfully attack different models. Liu~\etal~\cite{liu2016delving} then develop a stronger attack to successfully circumvent an online image classification system with ensemble attacks, which is later analysed by~\cite{liu2019deep}. Based on one of the most well-known attack methods, Fast Gradient Sign Method (FGSM)~\cite{goodfellow2014explaining} and its iteration-based version (I-FGSM)~\cite{kurakin2016adversarial}, many follow-ups are then proposed to further improve the transferability by adopting momentum term~\cite{dong2017boosting}, smoothing perturbation~\cite{Zhou_2018_ECCV}, constructing diverse inputs~\cite{xie2018improving}, augmenting ghost models~\cite{li2018learning} and smoothing gradient~\cite{dong2019evading}, respectively. Recent works \cite{baluja2017adversarial,mao2020gap++,naseer2019cross,poursaeed2019fine,poursaeed2017generative,xiao2018generating} also suggest to train generative models for creating adversarial examples. Besides transfer-based attacks, query-based~\cite{bhagoji2018practical,chen2017zoo,brendel2017decision,guo2018low,yang2020patchattack} attacks are also very popular black-box attack settings. % However, these families of methods need to access model outputs and query the target models for a large number of times. For example, Guo~\etal~\cite{guo2018low} limit the search space to a low-frequency domain and fool the Google Cloud Vision platform with an unprecedented 1000 model queries. %However, in security-sensitive scenarios, 1000 tries are still not practical.

\paragraph{Universal adversarial perturbations.}~Above are all image-dependent perturbation attacks. Moosavi-Dezfooli~\etal~\cite{moosavi2016universal} craft universal perturbations which can be directly added to any test images to fool the classifier with a high success rate. Poursaeed~\etal~\cite{poursaeed2017generative} propose to train a neural network for generating adversarial examples by explicitly feeding random noise to the network during training. After obtaining a well-trained model, they use a fixed input to generate universal adversarial perturbations. 
Researchers also explore to produce universal adversarial perturbations by different methods~\cite{Khrulkov_2018_CVPR,mopuri-bmvc-2017} or on different tasks~\cite{hendrik2017universal,poursaeed2017generative}. All these methods construct universal adversarial perturbations explicitly or data-independently. Unlike them, we provide an implicit data-driven alternative to generate universal adversarial perturbations.

\paragraph{Defense methods.}~Xie~\etal~\cite{xie2017mitigating} and Guo~\etal~\cite{guo2017countering} break transferability by applying input transformation such as random padding/resizing~\cite{xie2017mitigating}, JPEG compression~\cite{dziugaite2016study}, and total variance minimization~\cite{rudin1992nonlinear}. Injecting adversarial examples during training improves the robustness of deep neural network, termed as adversarial training. These adversarial examples can be pre-generated~\cite{liao2018defense,tramer2017ensemble} or generated on-the-fly during training~\cite{kannan2018adversarial,madry2017towards,xie2018feature}. Adversarial training is also applied to universal adversarial perturbations~\cite{akhtar2018defense,shafahi2018universal}.
% Tsipras~\etal~\cite{tsipras2018robustness} suggested, for an adversarially trained model, loss gradients in the input space align well with human perception.
% Shafahi~\etal~\cite{shafahi2018universal} also have a similar observation on the universal adversarially trained models.

\paragraph{Normalization.}~To induce regionally homogeneous perturbations, our work resorts to a new normalization strategy. This strategy appears similar to some normalization techniques, such as batch normalization~\cite{ioffe2015batch}, layer normalization~\cite{ba2016layer}, instance normalization~\cite{ulyanov2016instance}, group normalization~\cite{wu2018group},~\etc.~While these techniques aim to help the model converge faster and speed up the learning procedure for different tasks, the goal of our proposed region norm is to explicitly enforce the region structure and build homogeneity within regions.

\section{Regionally Homogeneous Perturbations}

As shown in Section~\ref{sec:intro}, regionally homogeneous perturbations appear to be strong in attacking defense models.
To acquire regionally homogeneous adversarial examples, we propose a gradient transformer module to generate regionally homogeneous perturbations from existing regionally non-homogeneous perturbations (\eg,~perturbations in the top row of Figure~\ref{fig:fig2}). In the following, we detail the transforming paradigm in Section~\ref{sec:framework} and the core component called gradient transformer module in Section~\ref{sec:region_homogeneity}, respectively. In Section~\ref{sec:universal_analysis}, we observe an under-fitting phenomenon and illustrate that the proposed gradient transformer module becomes quasi-input-independent, which benefits crafting universal adversarial perturbations. 

\subsection{Transforming Paradigm}\label{sec:framework}
To learn regionally homogeneous adversarial perturbations, we propose to use a shallow network $T$, which we call gradient transformer module, to transform the gradients that are generated by attacking naturally trained models.

Concretely, we consider Fast Gradient Sign Method (FGSM)~\cite{goodfellow2014explaining} which generates adversarial examples by
\begin{equation}\label{eq:FGSM}
I^{\text{adv}} = I + \epsilon\cdot\text{sign}\left(\nabla_{I} L(I, y; \theta)\right),
\end{equation}
where $L$ is the loss function of the model $\theta$, and sign$(\cdot)$ denotes the sign function. $y$ is the ground-truth label of the original image $I$. FGSM ensures that the generated adversarial example $I^{\text{adv}}$ is within the $\epsilon$-ball of $I$ in the $L_{\infty}$ space. 

Based on FGSM, we build pixel-wise connections via the additional gradient transformer module $T$, so that we may have regionally homogeneous perturbations. Therefore, Eqn.~\eqref{eq:FGSM} becomes
\begin{equation}
I^{\text{adv}} = I + \epsilon\cdot\text{sign}\left(T(\nabla_{I} L(I, y; \theta); \theta_T)\right),
\end{equation}
where $\theta_T$ is trainable parameter of gradient transformer module $T$, and we omit $\theta_T$ where possible for simplification.
The challenge we are facing now is how to train the gradient transformer module $T(\cdot)$ with the limited supervision. We address this by proposing a new transforming paradigm illustrated in Figure~\ref{fig:train1}. It consists of four steps, as we 1) compute the gradient $g=\nabla_{I} L(I, y; \theta)$ by attacking the naturally trained model $\theta$; 2) get the transformed gradient $\hat{g}=T(g; \theta_T)$ via the gradient transformer module; 3) construct the adversarial image $\hat{I}$ by adding the transformed perturbation to the clean image $I$, forward $\hat{I}$ to the same model $\theta$, and obtain the classification loss $L(\hat{I}^\text{adv}, y; \theta)$; and 4) freeze the clean image $I$ and the  model $\theta$, and update the parameters $\theta_T$ of $T(\cdot;\theta_T)$ by \textbf{maximizing} $L(\hat{I}^\text{adv}, y; \theta)$. The last step is implemented via stochastic gradient ascent (\eg,~we use the Adam optimizer~\cite{kingma2014adam} in our experiments). 

With the new transforming paradigm, one can potentially embed desirable properties via using the gradient transformer module $T(\cdot)$, and in the meantime, keep a high error rate on the model $\theta$. As we will show below, $T(\cdot)$ is customized to generate regionally homogeneous perturbations specially against defense models. Meanwhile, since we freeze the most part of the computation graph and leave a limited number of parameters (that is $\theta_T$) to optimize, the learning procedure is very fast.

\subsection{Gradient Transformer Module}\label{sec:region_homogeneity}
With the transforming paradigm aforementioned, we introduce the architecture of the core module, termed as gradient transformer module. The gradient transformer module aims at increasing the correlation of pixels in the same region, therefore inducing regionally homogeneous perturbations. As shown in Figure~\ref{fig:train2}, given a loss gradient $g$ as the input, the gradient transformer module $T(\cdot)$ is
\begin{equation}
    \hat{g} = T(g; \theta_T) = \mathrm{RN}\left(\mathrm{conv}(g)\right) + g,
\end{equation}
where $\mathrm{conv}(\cdot)$ is a $1 \times 1$ convolutional layer and $\mathrm{RN}(\cdot)$ is the newly proposed region norm layer. $\theta_T$ is the module parameters, which goes to the region norm layer ($\gamma$ and $\beta$ below) and the convolutional layer. A residual connection~\cite{he2016deep} is also incorporated. Since $\mathrm{RN(\cdot)}$ is initialized as zero~\cite{goyal2017accurate}, the residual connection allows us to insert the gradient transformer module into any gradient-based attack methods without breaking its initial behavior (\ie,~the transformed gradient $\hat{g}$ initially equals to $g$). Since the initial gradient $g$ is able to craft stronger adversarial example (compared with random noises), the gradient transformer module has a proper initialization.
The region norm layer consists of two parts, including a region split function and a region norm operator.

\vspace{1ex}\noindent\textbf{Region split function} splits an image (or equivalently, a convolutional feature map) into $K$ regions. Let $r(\cdot, \cdot)$ denote the region split function. The input of $r(\cdot,\cdot)$ is a pixel coordinate while the output is an index of the region which the pixel belongs to. With a region split function, we can get a partition $\{P_1, P_2, ..., P_K\}$ of an image, where $P_k = \{(h, w)\ |\ r(h, w) = k, 1\leq k\leq K\}$. 

\begin{figure}[tb]
  \begin{minipage}[c]{0.5\textwidth}
    \includegraphics[width=1\linewidth]{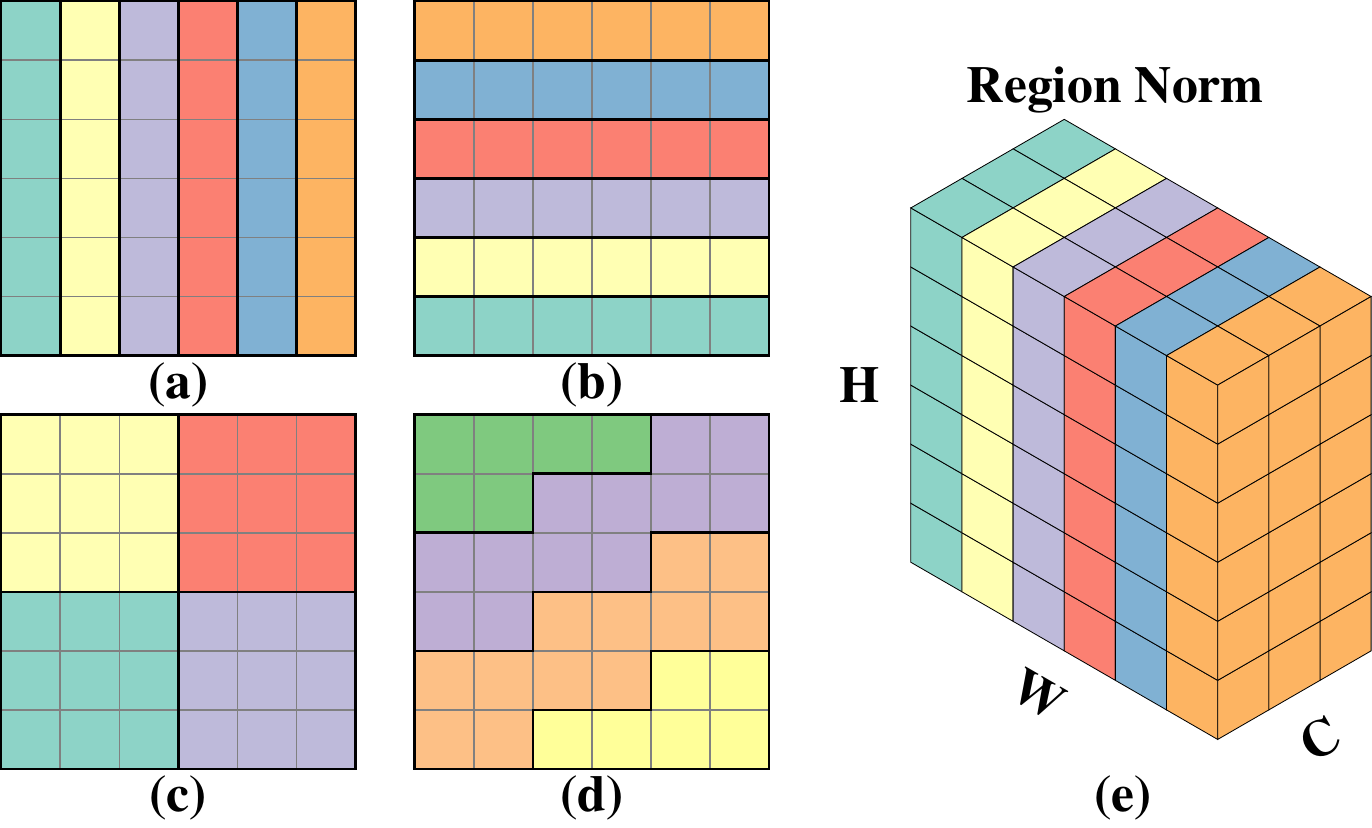}
  \end{minipage}\hfill
  \begin{minipage}[c]{0.47\textwidth}
    % \vspace{-1em}
    \caption{Toy examples of region split functions, including (a) vertical partition, (b) horizontal partition, (c) grid partition, and (d) slash partition. (e) illustrates the region norm operator with the region split function (a), where $C$ is the channel axis, $H$ and $W$ are the spatial axes. Each pixel indicates an $N$-dimensional vector, where $N$ is the batch size} \label{fig:splt_func}
  \end{minipage}
%   \vspace{-1em}
\end{figure}

In Figure~\ref{fig:splt_func}, we show 4 representatives of region split functions on a toy $6 \times 6$ image, including 1) vertical partition $(h, w) = w$, 2) horizontal partition $r(h, w) = h$, 3) grid partition $r(h, w) = \lfloor h / 3 \rfloor + 2\lfloor w / 3 \rfloor$, and 4) slash partition (parallel to an increasing line with the slope equal to 0.5).

\vspace{1ex}\noindent\textbf{Region norm operator} links pixels within the same region $P_k$, defined as
\begin{equation}\label{eq:norm}
y_i=\gamma_{k}\bar{x}_i+\beta_{k},\quad \bar{x}_i=\frac{1}{\sigma_{k}}(x_i-\mu_{k}),
\end{equation}
where $x_i$ and $y_i$ are the $i$-th input and output, respectively. And $i=(n, c, h, w)$ is a 4D vector indexing the features in $(N, C, H, W)$ order, where $N$ is the batch axis, $C$ is the channel axis, and $H$ and $W$ are the spatial height and width axes. We define $S_k$ as a set of pixels that belong to the region $P_k$, that is, $r(h,w)=k$.

$\mu_k$ and $\sigma_k$ in Eqn.~\eqref{eq:norm} are the mean and standard deviation (std) of the $k^\mathrm{th}$ region, computed by
\begin{equation}
\small
\begin{split}
 \mu_k&=\frac{1}{m_k}\sum_{j\in S_k}x_j, \\
 \sigma_k&=\sqrt{\frac{1}{m_k}\sum_{j\in S_k}(x_j-\mu_k)^2+\text{const}},
\end{split}
\end{equation}
where const is a small constant for numerical stability. $m_k$ is the size of $S_k$. Here $m_k=NC |P_k|$ and $|\cdot|$ is the cardinality of a given set. In the testing phase, the moving mean and moving std during training are used instead.
Since we split the image to regions, the trainable scale $\gamma$ and shift $\beta$ in Eqn.~\eqref{eq:norm} are also learned per-region. 

We illustrate the region norm operator in Figure~\ref{fig:splt_func}(e). To analyze the benefit, we compute the derivatives as
\begin{equation} \label{eq:bp_x}
\small
\begin{split}
&\frac{\partial L}{\partial \beta_k} = \sum_{j \in S_k} \frac{\partial L}{\partial y_j}, \quad
\frac{\partial L}{\partial \gamma_k} = \sum_{j \in S_k} \frac{\partial L}{\partial y_j}\bar{x}_j, \\
&\frac{\partial L}{\partial x_i} = \frac{1}{m_k\sigma_k}(m_k\frac{\partial L}{\partial \bar{x}_i} 
- \sum_{j \in S_k} \frac{\partial L}{\partial \bar{x}_j} 
- \bar{x}_i\sum_{j \in S_k} \frac{\partial L}{\partial\bar{x}_j}\bar{x}_j),
\end{split}
\end{equation}
where $L$ is the loss to optimize, and $\frac{\partial L}{\partial\bar{x}_i} = \frac{\partial L}{\partial y_i}\cdot\gamma_k$. It is not surprising that the gradient of $\gamma$ or $\beta$ is computed by all pixels in the related region. However, the gradient of a pixel with an index $i$ is also computed by all pixels in the same region. More significantly in Eqn.~\eqref{eq:bp_x}, the second term, $\sum_{j \in S_k} \frac{\partial L}{\partial \bar{x}_j},$ and the third term, $\bar{x}_i\sum_{j \in S_k} \frac{\partial L}{\partial\bar{x}_j}\bar{x}_j,$ are shared by all pixels in the same region. Therefore, the pixel-wise connections within the same region are much denser after inserting the region norm layer.

\paragraph{Comparison with other normalizations.}~Compared with existing normalizations (\eg,~Batch Norm~\cite{ioffe2015batch}, Layer Norm~\cite{ba2016layer}, Instance Norm~\cite{ulyanov2016instance} and Group Norm~\cite{wu2018group}), which aims to speed up learning, there are two main difference: 1) the goal of Region Norm is to generate regionally homogeneous perturbations, while existing methods mainly aim to stabilize and speed up training; 2) the formulation of Region Norm is splitting an image to regions and normalize each region individually, while other methods do not split along spatial dimension.

\subsection{Universal Analysis}\label{sec:universal_analysis}
By analyzing the magnitude of four probes ($a$, $b$, $c$, and $d$) in Figure~\ref{fig:train2}, we observe that $|b| >> |a|$ and $|c| >> |d|$ in a well-trained gradient transformer module (more results in Section~\ref{sec:universal_exp}). Consequently, such a well-trained module becomes quasi-input-independent,~\ie, the output is nearly fixed and less related to the input. Note that the output is still a little bit related to the input which is the reason why we use ``quasi-".

Here, we first build the connection between that observation and under-fitting to explain the reason. Then, we convert the quasi-input-independent module to an input-independent module for generating universal adversarial perturbations.

\paragraph{Under-fitting and the quasi-input-independent module.}
People figure out the trade-off between bias and variance of a model,~\ie,~the price for achieving a small bias is a large variance, and vice versa~\cite{bishop2006pattern,haykin2009neural}. Under-fitting occurs when the model shows low variance (but inevitable bias). An extremely low variance function gives a nearly fixed output whatever the input, which we term as quasi-input-independent.
Although in the most machine learning situation people do not expect this case, the quasi-input-independent function is desirable for generating universal adversarial perturbation.

Therefore, to encourage under-fitting, we go to the opposite direction of preventing under-fitting suggestions in~\cite{Goodfellow-et-al-2016}. On the one hand, to minimize the model capacity, our gradient transformer module only has $(12 + 2K)$ parameters, where a $3\times 3$ convolutional layer (bias enabled) incurs 12 parameters and $K$ is the number of region partitions. On the other hand, we use a large training data set $\mathcal{D}$ (5k images or more) so that the model capacity is relatively small. We then will have a quasi-input-independent module.

\paragraph{From quasi-input-independent to input-independent.}
According to the analysis above, we already have a quasi-input-independent module. To generate a universal adversarial perturbation, following the post-process strategy of Poursaeed~\etal~\cite{poursaeed2017generative}, we use a fixed vector as input of the module. Then following FGSM~\cite{goodfellow2014explaining}, the final universal perturbation will be $\Vec{u} = \epsilon\cdot\text{sign}(T(\Vec{z}))$, where $\Vec{z}$ is a fixed input. Recall that $\text{sign}(\cdot)$ denotes the sign function, and $T(\cdot)$ denotes the gradient transformer module.

\section{Experiments}
In this section, we demonstrate the effectiveness of the proposed regionally homogeneous perturbation (RHP) by attacking a series of defense models. The code is made publicly available.

\subsection{Experimental Setup}\label{sec:setup}
\paragraph{Dataset and evaluation metric.}~Without loss of generality, we randomly select $5000$ images from the ILSVRC 2012~\cite{deng2009imagenet} validation set to access the transferability of attack methods. For the evaluation metric, we use the improvement of top-$1$ error rate after attacking,~\ie,~the difference between the error rate of adversarial images and that of clean images.

\paragraph{Attack methods.}~For performance comparison, we reproduce five representative attack methods, including fast gradient sign method (FGSM)~\cite{goodfellow2014explaining}, momentum iterative fast gradient sign method (MIM)~\cite{dong2017boosting}, momentum diverse inputs iterative fast gradient sign method (DIM)~\cite{dong2017boosting,xie2018improving}, universal adversarial perturbations (UAP)~\cite{moosavi2016universal}, and the universal version of generative adversarial perturbations (GAP)~\cite{poursaeed2017generative}. % We will also compare our method with Transferable Adversarial Perturbations (TAP)~\cite{Zhou_2018_ECCV}. 
If not specified otherwise, we follow the default parameter setup in each method respectively.

To keep the perturbation quasi-imperceptible, we generate adversarial examples in the $\epsilon$-ball of original images in the $L_{\infty}$ space. The maximum perturbation $\epsilon$ is set as $16$ or $32$. The adversarial examples are generated by attacking a naturally trained network, Inception v3 (IncV3)~\cite{szegedy2016rethinking}, Inception v4 (IncV4) or Inception Resnet v2 (IncRes)~\cite{szegedy2017inception}. We use IncV3 and $\epsilon = 16$ in default.

\begin{table}[tb]
\small
\centering
\caption{The error rates (\%) of defense methods on our dataset which contains 5000 randomly selected ILSVRC 2012 validation images}
\label{table:clean-error-rate}
\begin{tabular}{|p{1.6cm}|*5{p{1.cm}<{\centering}|}p{1.4cm}<{\centering}|*3{p{1.cm}<{\centering}|}}
\hline
Defenses & TVM & HGD & R\&P & Inc\textsubscript{ens3} & Inc\textsubscript{ens4} & \small{IncRes\textsubscript{ens}} & PGD & ALP & FD  \\
\hline
Error Rate & 37.4 & 18.6 & 19.9 & 25.0 & 24.5 & 21.3 & 40.9 & 48.6 & 35.1 \\ \hline
\end{tabular}
\end{table}
%  We resize all ILSVRC 2012 validation images to $299 \times 299$ before test. 

\paragraph{Defense methods.}~As our method is to attack defense models, we reproduce nine defense methods for performance evaluation, including input transformation~\cite{guo2017countering} through total variance minimization (TVM), high-level representation guided denoiser (HGD)~\cite{liao2018defense}, input transformation through random resizing and padding (R\&P)~\cite{xie2017mitigating},
three ensemble adversarially trained models (Inc\textsubscript{ens3}, Inc\textsubscript{ens4} and IncRes\textsubscript{ens})~\cite{tramer2017ensemble}, adversarial training with project gradient descent white-box attacker (PGD)~\cite{madry2017towards,xie2018feature}, adversarial logits pairing (ALP)~\cite{kannan2018adversarial}, and feature denoising adversarially trained ResNeXt-101 (FD)~\cite{xie2018feature}. 

Among them, HGD~\cite{liao2018defense} and R\&P~\cite{xie2017mitigating} are the \emph{rank-1 submission} and \emph{rank-2 submission} in the NeurIPS 2017 defense competition~\cite{kurakin2018adversarial}, respectively. FD~\cite{xie2018feature} is the \emph{rank-1 submission} in the Competition on Adversarial Attacks and Defenses 2018. The top-1 error rates of these methods on our dataset are shown in Table~\ref{table:clean-error-rate}.

\paragraph{Implementation details.}~To train the gradient transformer module, we randomly select another $5000$ images from the validation set of ILSVRC 2012~\cite{deng2009imagenet} as the training set. Note that the training set and the testing set are disjoint.

For the region split function, we choose $r(h, w) = w$ as default, and will discuss different region split functions in Section~\ref{sec:region_split_results}.
% Following Cross-GPU Batch Norm~\cite{peng2018megdet}, the statistic values ($\mu$ and $\sigma$) of region norm layer are computed with a cross-GPU manner. 
We train the gradient transformer module for 50 epochs.
%with a total batch size 32 over 4 GPUs. 
When testing, we use a zero array as the input of the gradient transformer module to get universal adversarial perturbations,~\ie~the fixed input $\Vec{z} = \Vec{0}$. 

\begin{figure}[tb]
\centering
\includegraphics[width=0.97\linewidth]{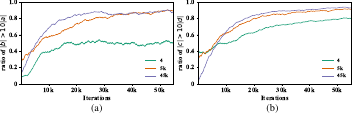}
% \vspace{-1em}
\caption{Universal analysis of RHP. In (a), we plot the ratio of the number of variables in probe pairs ($a$, $b$) satisfying that $|b|>10|a|$ to the total number of variables when training with 4, 5k or 45k images. In (b), we plot the case of $|c|>10|d|$}
\label{fig:underfit_curve}
% \vspace{-1em}
\end{figure}

\subsection{Under-fitting and Universal} \label{sec:universal_exp}
To verify the connections between under-fitting and universal adversarial perturbations, we change the number of training images so that the models are supposed to be under-fitting (due to the model capacity becomes low compared to large dataset)  or not. Specifically, we select 4, 5k or 45k images from the validation set of ILSVRC 2012 as the training set. We insert four probes $a$, $b$, $c$, and $d$ in the gradient transformer module as shown in Figure~\ref{fig:train2} and compare their values in Figure~\ref{fig:underfit_curve} with respect to the training iterations.

When the gradient transformer module is well trained with 5k or 45k images, we observe that: 1) $c$ overwhelms $d$, indicating the residual learning branch dominates the final output,~\ie,~$\hat{g} \approx c$; and 2) $b$ overwhelms $a$, indicating the output of the convolutional layer is less related to the input gradient $g$. Based on the two observations, we conclude that the gradient transformer module is quasi-input-independent when the module is under-fitted by a large number of training images in this case. Such a property is beneficial to generate universal adversarial perturbations (see Section~\ref{sec:universal_analysis}).

When the number of training images is limited (say 4 images), we observe that $b$ does not overwhelm $a$, indicating the output of the conv layer is related to the input gradient $g$, since a small training set cannot lead to under-fitting.

\begin{figure}[tb]
% \begin{center}
\includegraphics[width=0.9\linewidth]{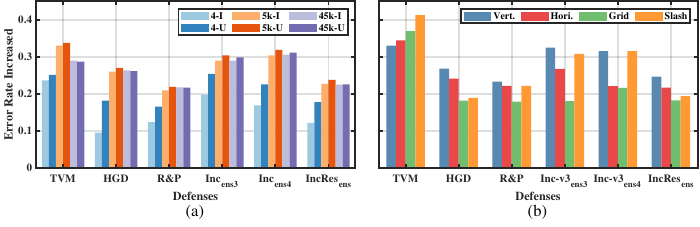}
% \end{center}
% \vspace{-1em}
\caption{(a) Performance comparison of universal (denoted by -U) inference and image dependent inference (denoted by -I) by varying the number of training images (4, 5k or 45k). (b) Performance comparison among four split functions, including vertical partition, horizontal partition, grid partition and slash partition}
\label{fig:underfit_split}
% \vspace{-1em}
\end{figure}

This conclusion is further supported by Figure~\ref{fig:underfit_split}(a): when training with 4 images, the performance gap between universal inference (use a fixed zero as the input of the gradient transformer module) and image dependent inference (use the loss gradient as the input) is quite large. The gap is reduced when using more data for training. 
% Figure~\ref{fig:underfit_split}(a) also illustrates 5k images are enough for under-fitting. Hence, we set the size of the training set to 5k in the following experiments.

To provide a better understanding of our implicit universal adversarial perturbation generating mechanism, we present an ablation study by comparing our method with other 3 strategies of generating universal adversarial perturbation with the same region split function. The compared includes 1) RP: Randomly assigns the Perturbation as $+\epsilon$ and $-\epsilon$ for each region; 2) OP: iteratively Optimizes the Perturbation to maximize classification loss on the naturally trained model (the idea of~\cite{moosavi2016universal}); 3) TU: explicitly Trains a Universal adversarial perturbations. The only difference between TU and our proposed RHP is that random noises take the place of the loss gradient $\textbf{g}$ in Figure~\ref{fig:train1} (following~\cite{poursaeed2017generative}) and are fed to the gradient transformer module. RHP is our proposed implicitly method, and the gradient transformer module becomes quasi-input-independent without taking random noise as the training input.

We evaluate above four settings on IncRes\textsubscript{ens}, the error rates increase by $14.0\%$, $19.4\%$, $19.3\%$, and $24.6\%$ for RP, OP, TU, and RHP respectively.
Since our implicit method has a proper initialization (Section~\ref{sec:region_homogeneity}), we observe that our implicit method constructs stronger universal adversarial perturbations.

\subsection{Transferability toward Defenses}\label{sec:main_results}
We first conduct the comparison in Table~\ref{table:main_results1} when the maximum perturbation $\epsilon=16$ and $32$, respectively. 

A first glance shows that compared with other representatives, the proposed RHP provides much stronger attack toward defenses. For example, when attacking HGD~\cite{liao2018defense} with $\epsilon=16$, RHP outperforms FGSM~\cite{goodfellow2014explaining} by $24.0\%$, MIM~\cite{dong2017boosting} by $19.5\%$, DIM~\cite{dong2017boosting,xie2018improving} by $14.9\%$, UAP~\cite{moosavi2016universal} by $24.9\%$, and GAP~\cite{poursaeed2017generative} by $25.5\%$, respectively. Second, universal methods generally perform worse than image-dependent methods as the latter can access and utilize the information from the clean images. Nevertheless, RHP, as a universal method, still beats those image-dependent methods by a large margin. 
At last, we observe that our method gains more when the maximum perturbation $\epsilon$ becomes larger.

\begin{table}[tb]
\small
\centering
\caption{The increase of error rates (\%) after attacking. The adversarial examples are generated with IncV3. In each cell, we show the results when the maximum perturbation $\epsilon=16/32$, respectively. The left 3 columns (FGSM, MIM and DIM) are image-dependent methods while the right 3 columns are (UAP, GAP and RHP) are universal methods}
\begin{tabular}{|c|ccc|ccc|}
\hline
Methods & FGSM~\cite{goodfellow2014explaining} & MIM~\cite{dong2017boosting} & DIM~\cite{dong2017boosting,xie2018improving} & UAP~\cite{moosavi2016universal} & GAP~\cite{poursaeed2017generative} & RHP (\textbf{ours}) \\ \hline
TVM & 21.9/45.3 & 18.2/37.1 & 21.9/41.0 & 4.78/12.1 & 18.5/50.1 & \textbf{33.0}/\textbf{56.9} \\ \hline
HGD & 2.84/20.7 & 7.30/18.7 & 11.9/32.1 & 1.94/11.3 & 1.34/37.9 & \textbf{26.8}/\textbf{57.5} \\ \hline
R\&P & 6.80/13.9 & 7.52/13.7 & 12.0/21.9 & 2.42/6.66 & 3.52/26.9 & \textbf{23.3}/\textbf{56.1} \\ \hline
Inc\textsubscript{ens3} & 10.0/17.9 & 11.4/17.3 & 16.7/26.1 & 1.00/7.82 & 5.48/33.3 & \textbf{32.5}/\textbf{60.8} \\ \hline
Inc\textsubscript{ens4} & 9.34/15.9 & 10.9/16.5 & 16.2/25.0 & 1.80/8.34 & 4.14/29.4 & \textbf{31.6}/\textbf{58.7} \\ \hline
IncRes\textsubscript{ens} & 6.86/13.3 & 7.76/13.6 & 10.8/19.6 & 1.88/5.60 & 3.76/22.5 & \textbf{24.6}/\textbf{57.0} \\ \hline
PGD & 1.90/12.8 & 1.36/6.86 & 1.84/7.70 & 0.04/1.04 & 1.28/10.2 & \textbf{2.40}/\textbf{25.8} \\ \hline
ALP & 17.0/32.3 & 15.3/24.4 & 15.5/24.7 & 7.98/11.5 & 15.6/30.0 & \textbf{17.8}/\textbf{39.4} \\ \hline
FD & 1.62/13.3 & 1.00/7.48 & 1.34/8.22 & -0.1/0.40 & 0.56/11.1 & \textbf{2.38}/\textbf{24.5} \\ \hline
\end{tabular}
\label{table:main_results1}
\end{table}

The performance comparison is also done when generating adversarial examples by attacking IncV4 or IncRes. Here we do not report the performance of GAP, because the official code does not support generating adversarial examples with IncV4 or IncRes. As shown in Table~\ref{table:main_results2} and Table~\ref{table:main_results3}, RHP still keeps strong against defense models. Meanwhile, it should be mentioned that when the model for generating adversarial perturbations is changed, RHP still generates universal adversarial examples. The only difference is that the gradients used in the training phase are changed, which then leads to a different set of parameters in the gradient transformer module.

\begin{table}[tb]
\centering
\caption{The increase of error rates (\%) after attacking. The adversarial examples are generated with IncV4. In each cell, we show the results when the maximum perturbation $\epsilon=16/32$, respectively. The left 3 columns (FGSM, MIM and DIM) are image-dependent methods while the right 2 columns are (UAP and RHP) are universal methods}
\begin{tabular}{|c|ccc|cc|}
\hline
Methods & FGSM~\cite{goodfellow2014explaining} & MIM~\cite{dong2017boosting} & DIM~\cite{dong2017boosting,xie2018improving} & UAP~\cite{moosavi2016universal} & RHP (\textbf{ours}) \\ \hline
TVM & 22.4/46.3 & 20.1/40.4 & 22.7/42.9 & 6.28/18.2 & \textbf{37.1}/\textbf{58.4} \\ \hline
HGD & 4.00/21.1 & 10.0/23.9 & 16.3/37.1 & 1.42/9.94 & \textbf{23.4}/\textbf{59.8} \\ \hline
R\&P & 8.68/15.1 & 10.2/17.4 & 14.7/25.0 & 2.42/6.52 & \textbf{20.2}/\textbf{57.6} \\ \hline
Inc\textsubscript{ens3} & 10.1/18.3 & 13.4/20.3 & 18.7/28.6 & 2.08/7.68 & \textbf{27.5}/\textbf{60.3} \\ \hline
Inc\textsubscript{ens4} & 9.72/17.4 & 13.1/19.0 & 17.9/26.5 & 1.94/6.92 & \textbf{26.7}/\textbf{62.5} \\ \hline
IncRes\textsubscript{ens} & 7.58/14.7 & 9.96/16.6 & 13.6/22.1 & 2.34/6.78 & \textbf{21.2}/\textbf{58.5} \\ \hline
PGD & 2.02/12.8 & 1.50/7.54 & 1.82/8.02 & 0.28/2.12 & \textbf{2.20}/\textbf{29.7} \\ \hline
ALP & 17.3/32.1 & 14.8/25.1 & 15.2/24.8 & 10.1/15.9 & \textbf{20.3}/\textbf{42.1} \\ \hline
FD & 1.42/13.4 & 1.24/8.18 & 1.62/8.74 & 0.16/1.18 & \textbf{1.90}/\textbf{31.8} \\  \hline
\end{tabular} 
\label{table:main_results2}
\end{table}

\begin{table}[tb]
\centering
\caption{The increase of error rates (\%) after attacking. The adversarial examples are generated with IncRes. In each cell, we show the results when the maximum perturbation $\epsilon=16/32$, respectively. The left 3 columns (FGSM, MIM and DIM) are image-dependent methods while the right 2 columns are (UAP and RHP) are universal methods}
\begin{tabular}{|c|ccc|cc|}
 \hline
Methods & FGSM~\cite{goodfellow2014explaining} & MIM~\cite{dong2017boosting} & DIM~\cite{dong2017boosting,xie2018improving} & UAP~\cite{moosavi2016universal} & RHP (\textbf{ours}) \\ \hline
TVM & 20.6/44.1 & 20.3/39.4 & 24.6/44.0 & 7.10/24.7 & \textbf{37.1}/\textbf{57.4} \\ \hline
HGD & 5.34/22.3 & 15.0/28.1 & 23.7/44.1 & 2.14/10.6 & \textbf{26.9}/\textbf{62.1} \\ \hline
R\&P & 10.1/15.8 & 13.4/22.1 & 22.5/34.5 & 2.50/8.36 & \textbf{25.1}/\textbf{61.4} \\ \hline
Inc\textsubscript{ens3} & 11.7/19.4 & 17.4/24.6 & 25.8/37.1 & 1.88/8.28 & \textbf{29.7}/\textbf{62.3} \\ \hline
Inc\textsubscript{ens4} & 10.5/17.2 & 15.1/22.5 & 22.4/33.7 & 1.74/7.22 & \textbf{29.8}/\textbf{63.3} \\ \hline
IncRes\textsubscript{ens} & 10.4/16.3 & 13.6/22.6 & 20.2/32.5 & 1.96/8.18 & \textbf{26.8}/\textbf{62.8} \\ \hline
PGD & 2.06/13.8 & 1.84/8.80 & \textbf{2.36}/9.26 & 0.40/3.78 & 2.20/\textbf{28.3} \\ \hline
ALP & 17.5/32.6 & 12.3/25.9 & 12.6/25.9 & 7.12/17.0 & \textbf{22.8}/\textbf{43.5} \\ \hline
FD & 1.72/14.7 & 1.62/9.48 & 1.78/10.1 & -0.1/3.06 & \textbf{2.20}/\textbf{32.2} \\ \hline
\end{tabular} 
\label{table:main_results3}
% \vspace{-0.25em}
\end{table}

\begin{figure}[tb]
\begin{center}
\includegraphics[width=0.9\linewidth]{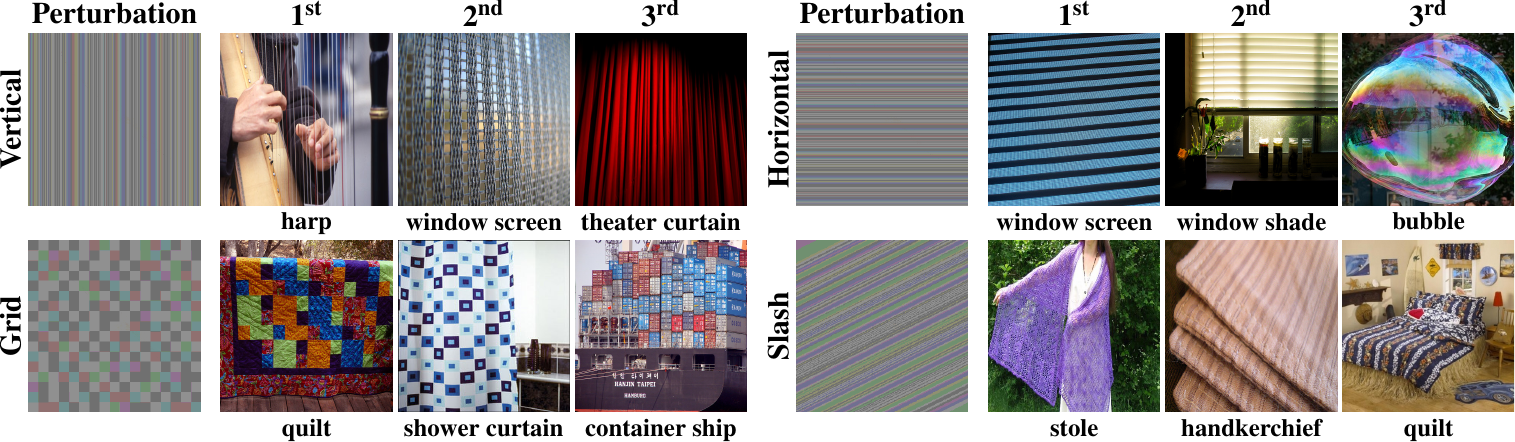}
\end{center}
% \vspace{-1em}
\caption{Four universal adversarial perturbations generated by different region split functions, and the corresponding top-3 target categories}
\label{fig:split_func_fig}
% \vspace{-1em}
\end{figure}

\begin{table}[tb]
\caption{Comparison of cross-task transferability. We attack segmentation model and test on the detection model Faster R-CNN, and report the value of mAP (lower is better for attacking methods). ``-" denotes the baseline performance without attacks}\label{table:cross-trans}
\centering
\begin{tabular}{|p{1.4cm}|*4{p{1.4cm}<{\centering}|}c|}
\hline
Attacks & - & FGSM & MIM   & DIM  & RHP (\textbf{ours}) \\ \hline 
mAP     & 69.2 & 43.1 & 41.6  & 36.2 & \textbf{31.6} \\ \hline
% map$\downarrow$ & 26.1 & 27.6 & 33.0 & \textbf{37.6} \\ \hline
\end{tabular}
% \vspace{-1em}
\end{table}

\subsection{Region Split Functions}\label{sec:region_split_results}
In this section, we discuss the choice of region split functions, \ie, vertical partition, horizontal partition, grid partition and slash partition (parallel to an increasing line with the slope equal to $0.5$).
Figure~\ref{fig:underfit_split}(b) shows the transferability to the defenses, which demonstrates that different region split functions are almost equivalently effective and all are stronger than our strongest baseline (DIM). Moreover, we observe an interesting phenomenon as presented in Figure~\ref{fig:split_func_fig}.

In each row of Figure~\ref{fig:split_func_fig}, we exhibit the universal adversarial perturbation generated by a certain kind of region split functions, followed by the top-3 categories to which the generated adversarial examples are most likely to be misclassified. For each category, we show a clean image as an exemplar. Note that our experiments are about the non-targeted attack, indicating the target class is undetermined and solely relies on the region split function.

As can be seen, the regionally homogeneous perturbations with different region split functions seem to be targeting at different categories, with an inherent connection between the low-level cues (\eg,~texture, shape) they share. For example, when using grid partition,  the top-3 target categories are quilt, shower curtain, and container ship, respectively, and one can observe that images in the three categories generally have grid-structured patterns.

Motivated by these qualitative results, we have a preliminary hypothesis that the regionally homogeneous perturbations tend to attack the low-level part of a model. The claim is not supported by a theoretical proof, however, it inspires us to test the cross-task transferability of RHP. 
As it is a common strategy to share the low-level CNN architecture/information in multi-task learning systems~\cite{ruder2017overview}, we conjecture that RHP can well transfer between different tasks (see below).

\subsection{Cross-task Transferability}\label{sec:x-task}

To demonstrate the cross-task transferability of RHP, we attack with the semantic segmentation task and test on the object detection task.

In more detail, we attack a semantic segmentation model (an Xception-65~\cite{chollet2017xception} based deeplab-v3+~\cite{deeplabv3plus2018}) on the Pascal VOC 2012 segmentation {\tt val}~\cite{everingham2015pascal}, and obtain the adversarial examples. Then, we take a VGG16~\cite{simonyan2015very} based Faster-RCNN model~\cite{Ren_2015_Faster}, trained on MS COCO~\cite{lin2014microsoft} and VOC2007 {\tt trainval}, as the testing model. To avoid testing images occurred in the training set of detection model, the testing set is the union of VOC2012 segmentation {\tt val} and VOC2012 detection {\tt trainval}, then we remove the images in VOC2007 dataset. The baseline performance of the clean images is mAP $69.2$. Here mAP score is the average of the precisions at different recall values.

As shown in Table~\ref{table:cross-trans}, RHP reports the lowest mAP with object detection, which demonstrates the stronger cross-task transferability than the baseline image-dependent perturbations,~\ie,~FGSM~\cite{goodfellow2014explaining}, MIM~\cite{dong2017boosting}, and DIM~\cite{dong2017boosting,xie2018improving}.

\section{Conclusion}
By white-box attacking naturally trained models and defense models, we observe the regional homogeneity of adversarial perturbations. Motivated by this observation, we propose a transforming paradigm and a gradient transformer module to generate the regionally homogeneous perturbation (RHP) specifically for attacking defenses. RHP possesses three merits, including 1) transferability: we demonstrate that RHP well transfers across different models (\ie,~transfer-based attack) and different tasks; 2) universal: taking advantage of the under-fitting of the gradient transformer module, RHP generates universal adversarial examples without explicitly enforcing the learning procedure towards it; 3) strong: RHP successfully attacks $9$ representative defenses and outperforms the state-of-the-art attacking methods by a large margin.

Recent studies~\cite{liao2018defense,xie2018feature} show that the mechanism of some defense models can be interpreted as a ``denoising" procedure. Since RHP is less like noise compared with other perturbations, it would be interesting to reveal the property of RHP from a denoising perspective in future works. Meanwhile, although evaluated with the non-targeted attack, RHP is supposed to be strong targeted attack as well, which requires further exploration and validation.

\section*{Acknowledgements}We thank Yuyin Zhou and Zhishuai Zhang for their insightful comments and suggestions. This work was partially supported by the Johns Hopkins University Institute for Assured Autonomy with grant IAA 80052272.

% \clearpage
\bibliographystyle{splncs04}
\bibliography{egbib}
\end{document}